\documentclass[runningheads]{llncs}

\usepackage{eccv}

\usepackage{eccvabbrv}

\usepackage{graphicx}
\usepackage{booktabs}

\usepackage[accsupp]{axessibility}  

\usepackage{hyperref}

\usepackage{orcidlink}

\begin{document}

\title{Growing Deep Neural Network Considering with Similarity between Neurons} 

\titlerunning{Abbreviated paper title}

\author{Taigo Sakai \inst{1}\orcidlink{0009-0000-5130-1522} \and
Kazuhiro Hotta\inst{1}\orcidlink{0000-0002-5675-8713}}

\authorrunning{T.Sakai et al.}

\institute{Meijo University, Nagoya, Japan \\
\email{200442066ccalumni.meijo-u.ac.jp,\\ Kazuhotta@meijo-u.ac.jp}}

\maketitle

\begin{abstract}
  Deep learning has excelled in image recognition tasks through neural networks inspired by the human brain. However, the necessity for large models to improve prediction accuracy introduces significant computational demands and extended training times. 
  Conventional methods such as fine-tuning, knowledge distillation, and pruning have the limitations like potential accuracy drops.
  Drawing inspiration from human neurogenesis, where neuron formation continues into adulthood, we explore a novel approach of progressively increasing neuron numbers in compact models during training phases, thereby managing computational costs effectively. We propose a method that reduces feature extraction biases and neuronal redundancy by introducing constraints based on neuron similarity distributions. This approach not only fosters efficient learning in new neurons but also enhances feature extraction relevancy for given tasks. Results on CIFAR-10 and CIFAR-100 datasets demonstrated accuracy improvement, and our method pays more attention to whole object to be classified in comparison with conventional method through Grad-CAM visualizations. These results suggest that our method's potential to decision-making processes.
\end{abstract}

\section{Introduction}
\label{intro}
\begin{figure}[t]
    \centering
    \begin{subfigure}[b]{0.55\textwidth}
        \centering
        \includegraphics[width=\textwidth]{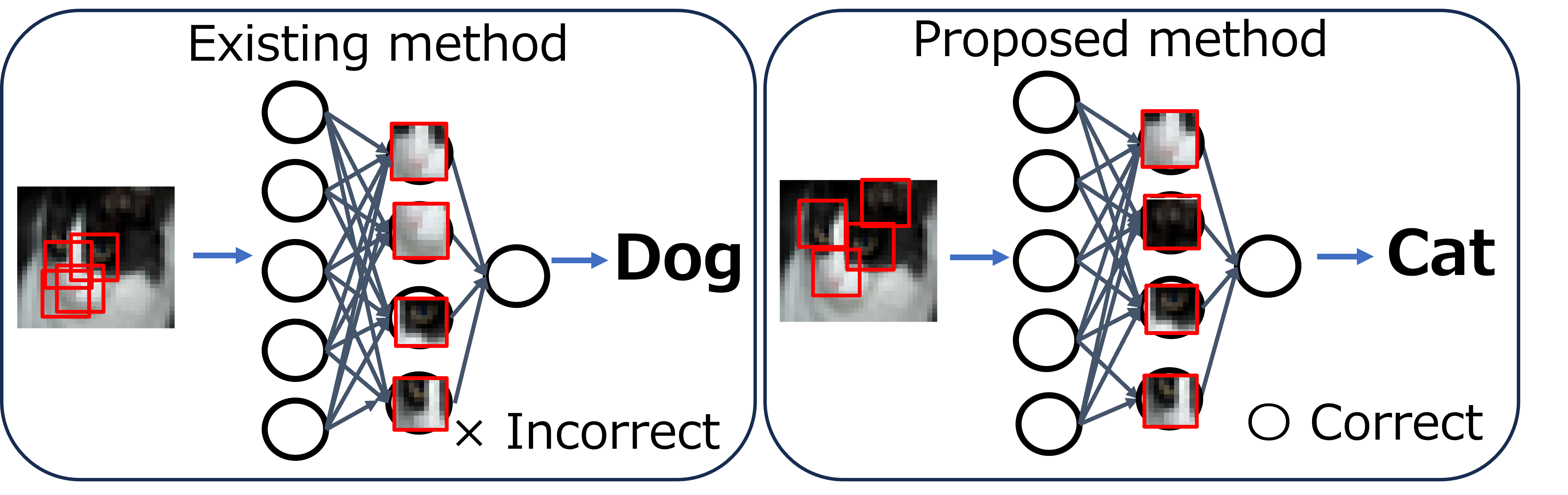}
        \caption{The conceptual diagram comparing the features extracted by existing methods and the proposed method.}
        \label{INTRO}
    \end{subfigure}
    \hfill
    \begin{subfigure}[b]{0.40\textwidth}
        \centering
        \begin{tabular}{@{}c@{\hspace{2mm}}c@{}}
            \includegraphics[width=0.48\textwidth]{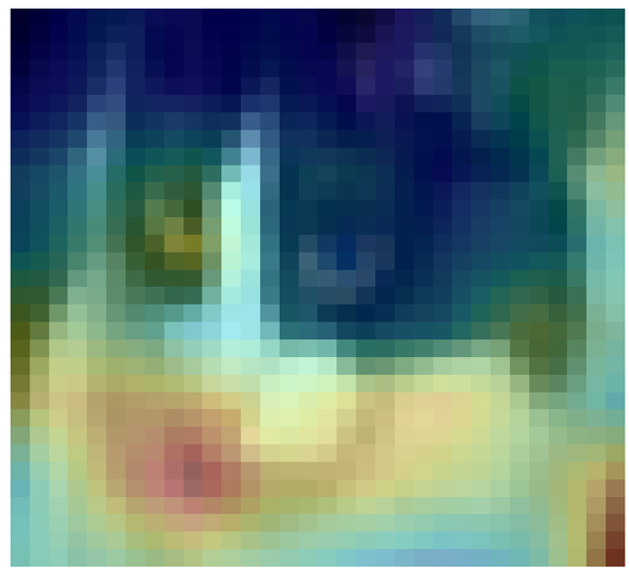} &
            \includegraphics[width=0.48\textwidth]{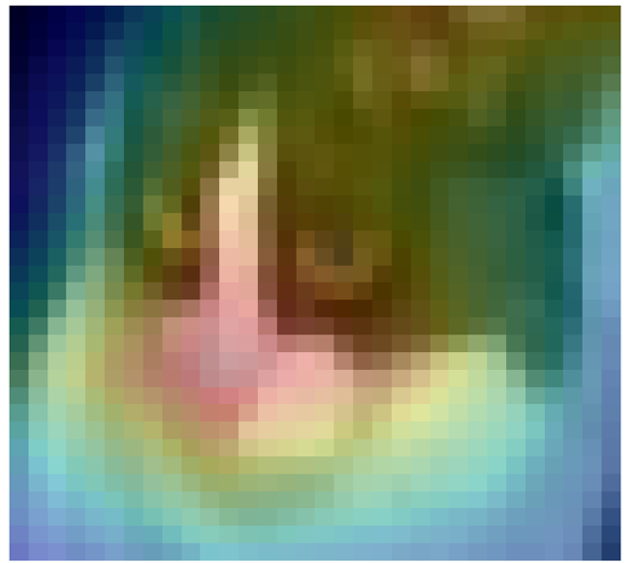}
        \end{tabular}
        \caption{Visualization results by Grad-CAM. Top: Input image. Bottom left: Firefly method. Bottom right: Our method.}
        \label{gradcamintro}
    \end{subfigure}
    \caption{Comparison of feature extraction and visualization results. (a) shows the comparison diagram of the existing method and our method. (b) shows the visualization using Grad-CAM. The left image represents the result from the conventional method (Firefly), while the right image shows the result from our proposed method.}
    \label{fig:combined}
\end{figure}

Deep learning, especially neural networks inspired by the human brain, has made remarkable progress in tasks like image classification\cite{NIPS2012c399862d}, segmentation\cite{RonnebergerFB15}, and generation\cite{goodfellow2014generative}. Larger models typically achieve higher accuracy by extracting rich features, but they require extensive computational resources and longer training times. To address these issues, researchers have proposed methods such as fine-tuning\cite{li2020rethinking}, knowledge distillation\cite{hinton2015distilling}, and pruning\cite{liu2019rethinking}. However, these approaches have drawbacks: fine-tuning may potential accuracy drops, knowledge distillation depends heavily on the original model's performance, and pruning can significantly degrade overall performance if not done carefully.

On the other hand, it is widely known that the process of neuron formation, or neurogenesis\cite{neurogenesis}, continues until adulthood in the human brain. From this background, the research that emulates brain neurogenesis by progressively increasing the number of neurons in small-scale networks during the training phases has gained attention \cite{liu2019splitting}, \cite{wu2021firefly}. Gradually enlarging models from compact sizes during training can help mitigate computational expenses. 
In the field of growing neural networks, the main methods of growing include the random addition of new neurons and the duplication of existing neurons. However, Both approaches induce the redundancy among neurons within the network. Random additions can lead to uneven distributions of neurons, resulting in a tendency to over-extract certain features. Similarly, the duplication of neurons lead to biases in feature extraction. For simplicity, a comparison diagram of the existing method and our method is shown in \cref{INTRO}. 

Existing methods often add new neurons that behave similarly to old ones (see left side of \cref{INTRO}). This focuses on the same image parts (e.g., a cat's eyes) in more detail, without learning new information or improving accuracy.
In contrast, our proposed method (right side of \cref{INTRO}) ensures new neurons behave differently from existing ones. This allows the network to notice previously overlooked image parts (e.g., a cat's ears or mouth), capturing a wider range of features and improving the classification accuracy.
Grad-CAM based visualization in \cref{gradcamintro} further illustrates this concept. Firefly (left side image) concentrates on specific image areas, while our method (right side image) demonstrates a more comprehensive feature extraction approach, attending to a broader range of relevant areas. This leads to a more accurate classification results.

This paper proposes a novel approach to address feature biases caused by neuronal redundancy by introducing constraints based on the distribution of similarities among neurons. Our method improves overall classification accuracy on CIFAR-10 and CIFAR-100 datasets. As demonstrated through Grad-CAM visualizations, our method enables the network to focus on the relevant parts of the target object rather than irrelevant background features. This comprehensive approach not only enhances performance but also mitigates shortcut learning\cite{onpro}, a common issue in continual learning where networks rely on superficial features. Our method improves the network's ability to identify key features of the target objects, leading to more accurate and reliable classifications within the tested image classification tasks.

This paper is organized as follows. \cref{Related} describes to related works. \cref{Proposed} explains the details of the proposed method. \cref{Experiments} presents experimental results to validate the efficacy of the proposed method and to compare with conventional method. 
\cref{Conclusion} describes the conclusions and prospects for future research.

\section{Related Works}
\label{Related}

\begin{figure}[h]
    \begin{center}
    \includegraphics[scale=0.48]{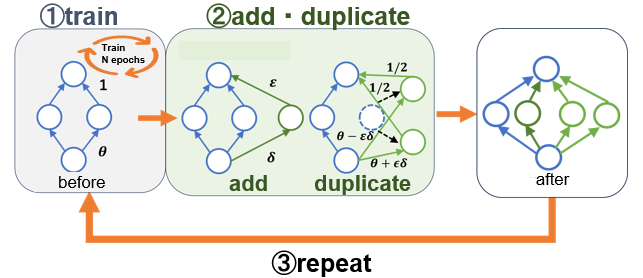}
    \end{center}
    \caption{Overall workflow of growing neurons algorithm.}
    \label{GROWFLOW}
\end{figure}

This section explains conventional growing neural networks, focusing on two main approaches: adding new neurons\cite{Dota2, chen2016net2net} and splitting existing neurons\cite{liu2019splitting}.
\cref{GROWFLOW} illustrates the overall workflow of growing neurons algorithms, consisting of two strategies: (1) duplicating existing neurons, or (2) adding new neurons with random weights.

\textbf{Random method:} Methods like Dota2\cite{Dota2} and Net2Net\cite{chen2016net2net} add randomly initialized neurons. While this rapidly expands network capacity, it risks inconsistencies with existing knowledge and inefficient learning.

\textbf{Splitting method:} The Splitting Steepest Descent (SSD) method\cite{liu2019splitting} duplicates existing neurons, stabilizing the learning process. However, this can lead to redundancy and prevent from learning new information.

\textbf{Firefly:\cite{wu2021firefly}} This algorithm combines both approaches, but fails to consider distribution relationships among neurons, leading to redundancy.

Our study proposes the approach to use similarity between neurons to reduce redundancy, improve efficiency, and prevent shortcut learning. We introduce post-neuron increment processes:
\begin{enumerate}
    \item[(a)] Compute similarity between neurons and create a similarity map
    \item[(b)] Flatten the map into a one-dimensional distribution
    \item[(c)] Guide learning to control the average cosine similarity distribution towards zero
\end{enumerate}

This novel approach optimizes the entire set of neurons when adding new ones, potentially preventing biased learning by enabling neurons to assume distinct roles.

\section{Proposed Method}
\label{Proposed}

\begin{figure}[h]
    \centering
    \begin{tabular}{cc}
      \begin{minipage}[t]{0.55\textwidth}
        \centering
        \includegraphics[width=\textwidth]{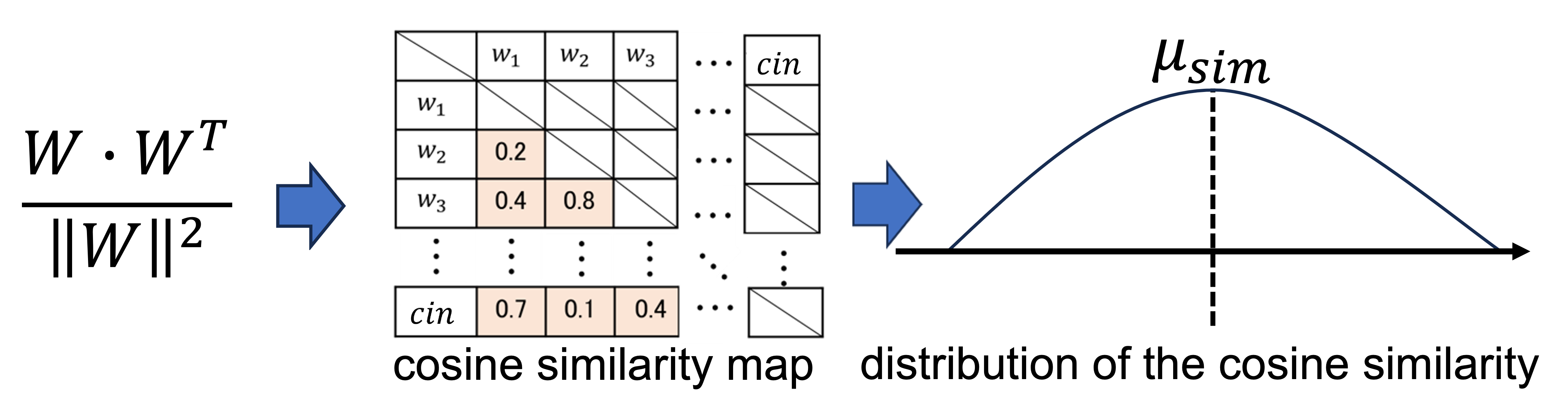}
        \subcaption{The computational flow of our method.}
        \label{METHOD2}
      \end{minipage}
      &
      \begin{minipage}[t]{0.41\textwidth}
        \centering
        \includegraphics[width=\textwidth]{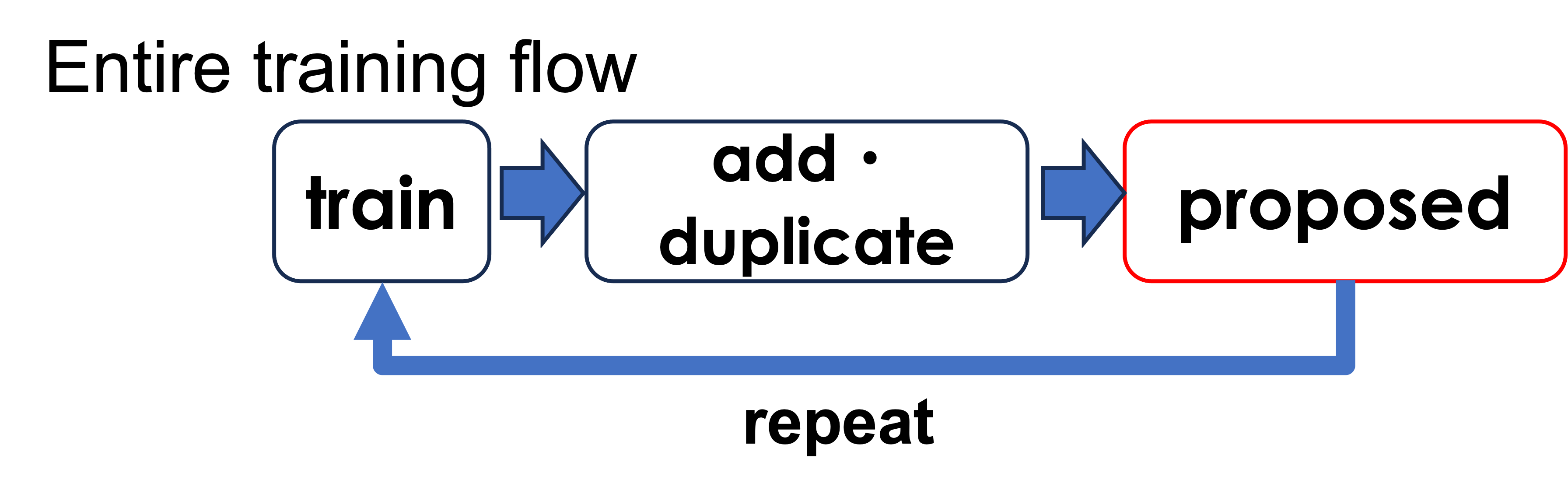}
        \subcaption{The entire training flow in our method.}
        \label{METHOD1}
      \end{minipage}
    \end{tabular}
    \caption{The overview of the proposed method.}
\end{figure}

\cref{METHOD2} and \cref{METHOD1} 
illustrate the overview of the proposed method. W represents the number of input channels \(C_{\text{in}}\) and output channels \(C_{\text{out}}\) of a neuron, and the neuron matrix \(W\) belongs to \(\mathbb{R}^{C_{\text{out}} \times C_{\text{in}}}\). Although we explain the details later, the overall process involves the computation of transpose product of W to obtain a cosine similarity map between neurons with dimensions \(C_{\text{out}} \times C_{\text{out}}\). Minimizing cosine similarity aims to give each neuron a different "specialization", enabling efficient distribution of distinct roles rather than duplicating the same task, thus acquiring more diverse features.
The algorithm trains to ensure that the average value of the cosine similarity distribution is close to zero.
This algorithm is performed after the addition and duplication phase shown in the related work. Existing techniques like SSD and Firefly involve duplication of existing neurons and addition of new neurons. However, duplication of existing neurons does not facilitate the extraction of new information. Additionally, the approach that adds new neurons randomly does not take into account the relationships with existing neurons, leading to prolonged convergence times and no guarantee that the new neurons will assume novel roles.
This paper proposes a method that ensures each neuron assumes a distinct role, enabling the effective extraction of task-relevant features. This is achieved by distributing the cosine similarities between dimensions in each layer of the neural network and make the average of the cosine similarity distribution closer to 0.
The method primarily consists of two steps: computation of the cosine similarity map ( \cref{calcosine}) and minimization of the mean of cosine similarities ( \cref{MU,minmu}).


\subsection{Computation and Adjustment of Cosine Similarity}
\label{cosine_and_adjustment}

We compute the cosine similarity map \(C_{\text{sim}}\) between neurons using the normalized weight matrix \(W_{\text{norm}}\):
\begin{equation}\label{calcosine}
   C_{\text{sim}} = W_{\text{norm}} \cdot W_{\text{norm}}^T, \quad \text{where} \quad W_{\text{norm}} = \frac{W}{\|W\|}.
\end{equation}
The mean \(\mu_{\text{sim}}\) of the similarity map, excluding diagonal elements, is calculated as:
\begin{equation}\label{MU}
\mu_{\text{sim}} = \frac{1}{n(n-1)} \sum_{i=1}^{n} \sum_{j=1, j \neq i}^{n} |(C_{\text{sim}}(i, j) - I)|
\end{equation}
where \(n\) is the matrix dimension. Our objective is to minimize \(\mu_{\text{sim}}\):
\begin{equation}\label{minmu}
\min Loss = \min \mu_{\text{sim}}.
\end{equation}
This approach aims to distribute neuron similarities more evenly, potentially reducing redundancy and improving feature extraction.

\subsection{Prevent changes in weights}
\label{preventchangesweights}
In this paper, we introduce a specific loss term to suppress unstable changes in the sum of weights by constraint Eq. (\ref{MU}). Specifically, by incorporating the change in weights of each layer into the loss function using the formula as follows:
\begin{equation}
\label{currentpreviousweght}
\log \left( \left| \frac{\sum_{i=1}^{C_{\text{out}}} \sum_{j=1}^{C_{\text{in}}} W_{\text{current}}}{\sum_{i=1}^{C_{\text{out}}} \sum_{j=1}^{C_{\text{in}}} W_{\text{previous}}}\right| \right)
\end{equation}
where $W_{\text{current}}, W_{\text{previous}}$ indicates current, previous weight matrix, as $W_{\text{current}}, W_{\text{previous}} \in \mathbb{R}^{C_{\text{out}} \times C_{\text{in}}}$.

We aim to avoid sudden increases or decreases in weights.
Finally, the overall objective function is to minimize the sum of Eq. (\ref{MU}) and Eq. (\ref{currentpreviousweght}), written as:
\begin{equation}
\label{overallfunc}
\min Loss = \min \left\{\mu_{\text{sim}} + \lambda \times \log \left(\frac{\sum_{i=1}^{C_{\text{out}}} \sum_{j=1}^{C_{\text{in}}} W_{\text{current}}}{\sum_{i=1}^{C_{\text{out}}} \sum_{j=1}^{C_{\text{in}}} W_{\text{previous}}}\right)\right\}
\end{equation}
where $\lambda$ is an arbitrary constant, and we configure different $\lambda$ in experiments.

\section{Experiments}
\label{Experiments}
\subsection{Instrumentation and Equipment}

We evaluated our method against Random growing, SSD, and Firefly on image classification tasks using CIFAR-10 and CIFAR-100 datasets\cite{CIFAR}. We employed VGG16\cite{vgg} and MobileNetv1\cite{mobilenet} architectures, training for 250 epochs using SGD\cite{sgd} with a momentum of 0.9 and cosine learning rate decay. Networks were grown by 35\% of their current parameter size every 50 epochs. 
The number of initial and final parameters was 0.0288 million and 0.325 millions, respectively. Experiments were conducted using NVIDIA GTX1080Ti GPUs with a batch size of 128. Detailed network configurations followed those in SSD and Firefly studies, with intermediate layer channel sizes set to 16 and 32.

\subsection{Experimental results}
\label{resultacc}

Tables \ref{AveAcc10} and \ref{AveAcc100} show the experimental results. $N_{best}$ indicates the optimal number of iterations that yielded the highest accuracy, chosen from experiments with $N$ varying from 1 to 30. Our proposed method consistently improved the accuracy across Random, SSD, and Firefly algorithms with only a slight increase in computational cost. The optimal $N_{best}$ varied among the different methods, suggesting the need for method-specific tuning.

\begin{table}[ht]
\centering
\begin{minipage}[t]{0.48\textwidth}
    \centering
    \caption{Average test accuracy and training times on CIFAR-10 dataset.}
    \label{AveAcc10}
    \scalebox{0.6}{
        \begin{tabular}{l|l|c|c|c}
        \hline
        \textbf{Method}       & \textbf{Architecture} & \multicolumn{1}{l|}{\textbf{Accuracy(\%)}} & \multicolumn{1}{l|}{\textbf{Training time(s)}} & \multicolumn{1}{l}{\ \ \ $N_{best}$ \ \ \ } \\ \hline
        \textbf{Random}       & \textbf{VGG16}        & $86.49$& 3022& -\\ \cline{2-5} 
        \textbf{}             & \textbf{MobileNet}    & $87.90$& 1240& -\\ \hline
        \textbf{Random}       & \textbf{VGG16}        & $86.85$& 3090& 20\\ \cline{2-5} 
        \textbf{+ours}        & \textbf{MobileNet}    & $88.29$& 1264& 20\\ \hline
        \textbf{SSD}          & \textbf{VGG16}        & $89.12$& 6146& -\\ \cline{2-5} 
        \textbf{}             & \textbf{MobileNet}    & $89.92$& 3191& -\\ \hline
        \textbf{SSD}          & \textbf{VGG16}        & $90.08$& 6176& 10\\ \cline{2-5} 
        \textbf{+ours}        & \textbf{MobileNet}    & $90.83$& 3370& 10\\ \hline
        \textbf{Firefly}      & \textbf{VGG16}        & $90.60$& 3375& -\\ \cline{2-5} 
        \textbf{}             & \textbf{MobileNet}    & $91.89$& 1630& -\\ \hline
        \textbf{Firefly}      & \textbf{VGG16}        & $91.70$& 3409& 15\\ \cline{2-5} 
        \textbf{+ours}        & \textbf{MobileNet}    & \textcolor{red}{$92.56$}& 1670& 15\\ \hline
        \end{tabular}
    }
\end{minipage}
\hfill
\begin{minipage}[t]{0.48\textwidth}
    \centering
    \caption{Average test accuracy and training times on CIFAR-100 dataset.}
    \label{AveAcc100}
    \scalebox{0.6}{
        \begin{tabular}{l|l|c|c|c}
        \hline
        \textbf{Method}       & \textbf{Architecture} & \multicolumn{1}{l|}{\textbf{Accuracy(\%)}} & \multicolumn{1}{l|}{\textbf{Training time(s)}} & \multicolumn{1}{l}{\ \ \ $N_{best}$ \ \ \ } \\ \hline
        \textbf{Random}       & \textbf{VGG16}        & $61.77$& 6612& -\\ \cline{2-5} 
        \textbf{}             & \textbf{MobileNet}    & $62.89$& 2690& -\\ \hline
        \textbf{Random}       & \textbf{VGG16}        & $62.09$& 6680& 20\\ \cline{2-5} 
        \textbf{+ours}        & \textbf{MobileNet}    & $63.10$& 2842& 20\\ \hline
        \textbf{SSD}          & \textbf{VGG16}        & $63.90$& 13560& -\\ \cline{2-5} 
        \textbf{}             & \textbf{MobileNet}    & $65.71$& 7167& -\\ \hline
        \textbf{SSD}          & \textbf{VGG16}        & $64.52$& 13790& 10\\ \cline{2-5} 
        \textbf{+ours}        & \textbf{MobileNet}    & $66.16$& 7250& 10\\ \hline
        \textbf{Firefly}      & \textbf{VGG16}        & $64.74$& 5307& -\\ \cline{2-5} 
        \textbf{}             & \textbf{MobileNet}    & $66.42$& 3673& -\\ \hline
        \textbf{Firefly}      & \textbf{VGG16}        & $65.55$& 5360& 15\\ \cline{2-5} 
        \textbf{+ours}        & \textbf{MobileNet}    & \textcolor{red}{$67.49$}& 3690& 15\\ \hline
        \end{tabular}
    }
\end{minipage}
\end{table}

\begin{figure}[h]
\begin{minipage}[c]{0.45\vsize}
    \scalebox{0.6}{
        \begin{tabular}{c}
            \includegraphics[scale=0.68]{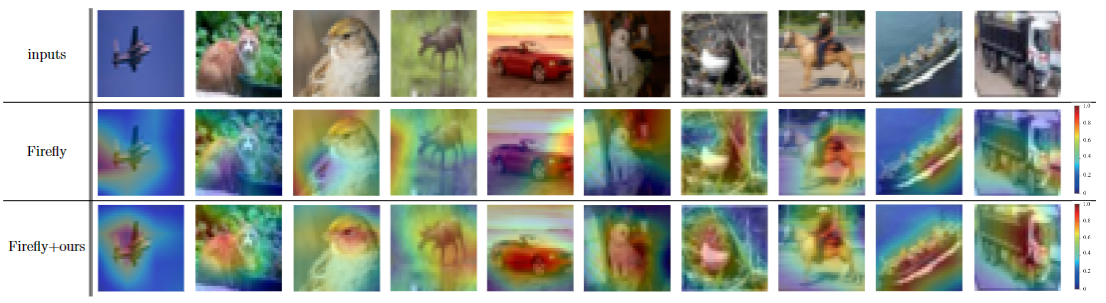} \\
        \end{tabular}
    }
\end{minipage}
\caption{Grad-CAM visualization results of VGG16 on CIFAR-10 dataset. Columns show different classes. Top: inputs; Middle: Firefly; Bottom: Firefly+ours.}
    \label{grad10}

\vfill
\begin{minipage}[c]{0.45\vsize}
    \scalebox{0.55}{
        \begin{tabular}{c}
            \includegraphics[scale=0.68]{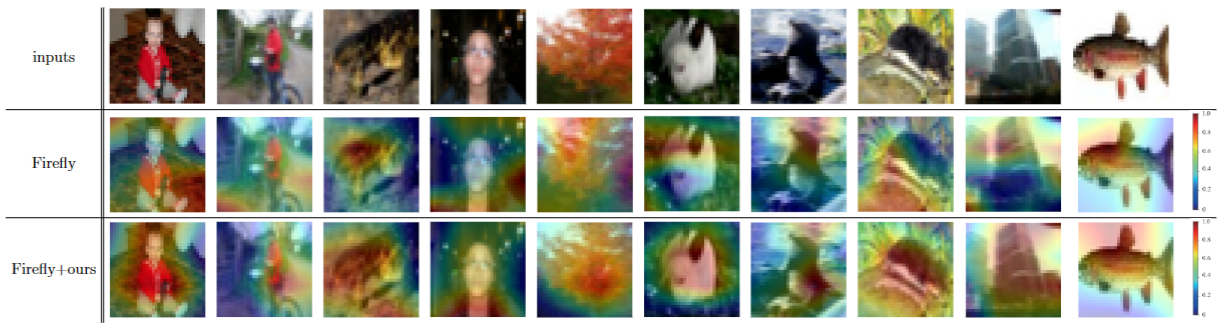} \\
        \end{tabular}
    }
    \end{minipage}
\caption{Grad-CAM visualization results of VGG16 on CIFAR-100 dataset. Columns show different classes. Top: inputs; Middle: Firefly; Bottom: Firefly+ours.}
    \label{grad100}
\end{figure}

We used Grad-CAM to visualize the network's focus areas, and the results are shown in Figures \ref{grad10} and \ref{grad100}. The Firefly model often focused on irrelevant areas or specific object parts, potentially leading to shortcut learning\cite{onpro}. In contrast, our method distributed attention more evenly across relevant object areas. This improved focus on overall features likely contributes to the enhanced test accuracy observed with our approach.

\begin{figure}[h]
\centering
\begin{minipage}[b]{0.48\linewidth}
    \centering
    \includegraphics[scale=0.29]{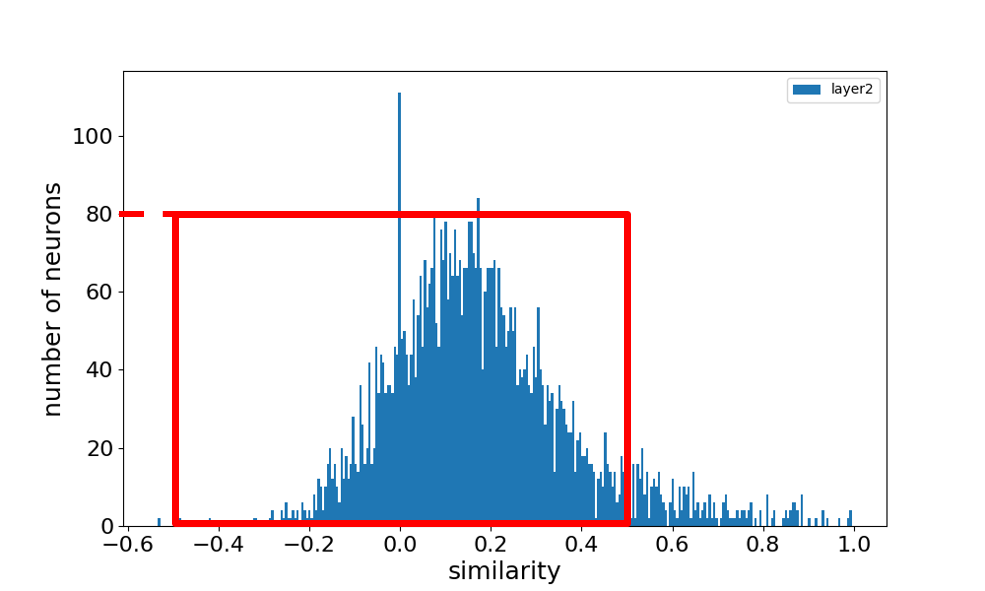}
    \subcaption{Firefly without our method}
    \label{distnone}
\end{minipage}
\begin{minipage}[b]{0.48\linewidth}
    \centering
    \includegraphics[scale=0.29]{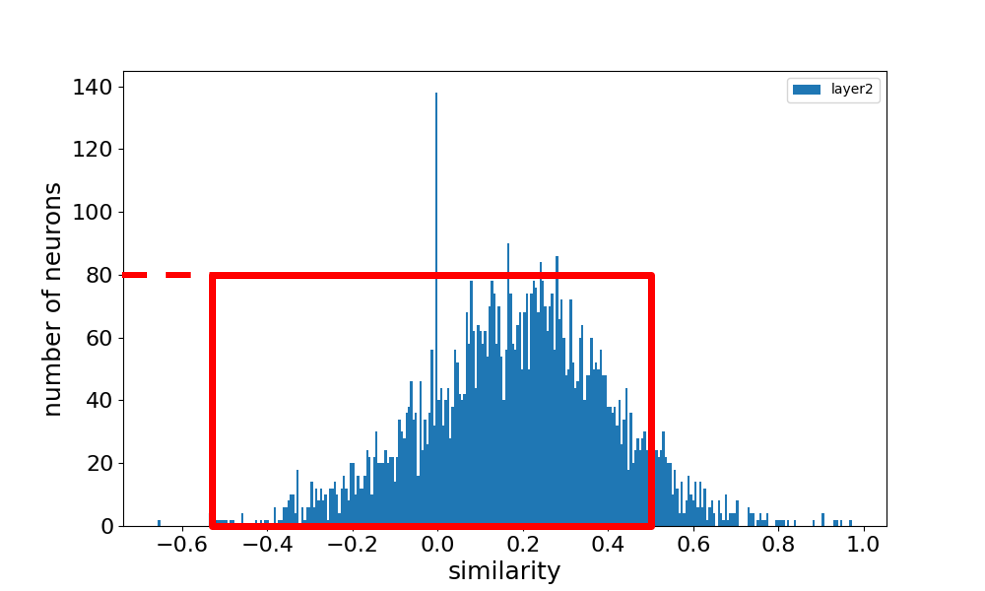}
    \subcaption{Firefly with our method}
    \label{distcos}
    
\end{minipage}
\caption{Distribution for cosine similarlity in Firefly. The vertical axis represents the number of neuron pairs, while the horizontal axis shows their cosine similarity. (a) represents the scenario without our method, and (b) shows the results with our method. The area enclosed by the red frame indicates the range where the cosine similarity is between -0.5 and 0.5, highlighting the region close to 0.}
\label{cossimdist}
\end{figure}

In this paper, we propose an approach that aims to bring the average absolute value of the cosine similarity distribution closer to 0, as presented in \cref{minmu}. To demonstrate this approach, we visualized the distribution of cosine similarity after training and compared scenarios with and without our proposed method. \cref{cossimdist} shows the plot results of the cosine similarity distribution in the second layer of VGG16 with the Firefly method. We focused on Layer 2 because diverse and effective learning at this layer enables more complex feature extraction in subsequent layers. Additionally, Layer 2 is optimally positioned to observe both the influence of input data and its impact on subsequent layers, allowing for clear evaluation of the proposed method's effectiveness. The reason of selecting the second layer is to verify whether the constraints of the proposed method are reflected in the shallower layers of the model. The first layer was not selected due to its fixed number of input channels, limited number of neurons, and the variability in distribution observed across experiments. 
The results shows that the scenario with our method \cref{distcos} has a greater number of neuron pairs closer to 0 compared to the scenario without our method. 
We believe that we can control the distribution to be more evenly centered around 0.

Furthermore, by aiming for an average closer to 0, we hypothesize that it is possible to prevent an increase in neurons that are extremely similar or dissimilar, while enabling a more effective distribution of roles in extracting diverse information.

\subsection{Ablation Study}

We evaluate the model's accuracy across three distinct scenarios: First, the method without the application of minimizing absolute value of cosine similality \cref{minmu}(temporary called Proposed1) and preventing the change of sum of weights \cref{preventchangesweights} (Proposed2). 
This represents the conventional method. 
Second, the method with only Proposed1. Third, 
the method with only Proposed2. Fourth, the method with both Proposed1 and Proposed2 together. The result is as shown in \cref{withoutabs}.

The result indicates that the combination of techniques Proposed1 and Proposed2 yields the most improvement in accuracy. On the other hand, applying only technique Proposed2 resulted in accuracy comparable to the conventional method. 
This suggests that the application of only Proposed1 results in slight improvement in accuracy, because due to altering the sum of weights,
The Proposed2 mitigated this effect, leading to an overall enhancement in accuracy.

\begin{table}[h]
    \centering
    \caption{Comparison result of average test accuracy and training times on CIFAR-10 dataset. Proposed1 indicates minimization of absolute mean of cosine similality, while Proposed2 is preventing the altering the sum of weights.}
    \label{withoutabs}
    \scalebox{0.9}{
        \begin{tabular}{l|cc|c|c|c}
        \hline
        \textbf{Method}       & \textbf{Proposed1} & \textbf{Proposed2} & \multicolumn{1}{l|}{\textbf{Accuracy(\%)}} & \multicolumn{1}{l|}{\textbf{Training time(s)}} & \multicolumn{1}{l}{\ \ \ $N_{best}$ \ \ \ } \\ \hline
        \textbf{Random}       & \textbf{$\times$}& \textbf{$\times$}        & $86.49$& 3022& -\\ \hline
        \textbf{Random+ours}  & \textbf{$\times$} & \textbf{$\circ$}        & $86.49$& 3030& 20\\ \cline{2-6} & \textbf{$\circ$} & \textbf{$\times$}
        & $86.67$& 3077& 20\\ \cline{2-6} & \textbf{$\circ$} & \textbf{$\circ$}
        & $\mathbf{86.85}$& 3090& 20\\ \hline 
        \textbf{SSD}          & \textbf{$\times$}& \textbf{$\times$}        & $89.12$& 6146& -\\ \hline
        \textbf{SSD+ours}     & \textbf{$\times$} & \textbf{$\circ$}        & $89.12$& 6149& 10\\ \cline{2-6}
        \textbf{}             & \textbf{$\circ$} & \textbf{$\times$}    & $89.58$& 6171& 10\\  \cline{2-6} & \textbf{$\circ$} & \textbf{$\circ$}
        & $\mathbf{90.08}$& 6176& 10\\ \hline 
        \textbf{Firefly}      & \textbf{$\times$}& \textbf{$\times$}        & $90.60$& 3375& -\\ \hline 
        \textbf{Firefly+ours} & \textbf{$\times$} & \textbf{$\circ$}        & $91.60$& 3379& 15\\ \cline{2-6} 
        \textbf{}             & \textbf{$\circ$} & \textbf{$\times$}    & $91.51$& 3402& 15\\ \cline{2-6} & \textbf{$\circ$} & \textbf{$\circ$}
        & $\mathbf{91.70}$& 3409& 15\\ \hline 
        \end{tabular}
        }
\end{table}


\section{Conclusion}
\label{Conclusion}
In this paper, we propose a new method to consider the similarity between neurons and train neural networks so that the average of the similarity distribution approaches 0 in the learning process of neural networks which grow during training.
Experimental results show that our method allows the network to learn more diverse features, increasing the overall complexity and discrimination ability. Quantitative comparative experiments on image classification tasks using CIFAR-10 and CIFAR-100 datasets and qualitative analysis using Grad-CAM showed that the proposed method enables the network to concentrate on the important area for target classification and become capable of inferring by observing the target, rather than making hasty judgments based on partial views, similar to human perception.

\section*{Acknowledgments}
This work was partially supported by JSPS KAKENHI Grant Number 24K15020.


%
%
\bibliographystyle{splncs04}
\bibliography{main}

\begin{thebibliography}{10}
\providecommand{\url}[1]{\texttt{#1}}
\providecommand{\urlprefix}{URL }
\providecommand{\doi}[1]{https://doi.org/#1}

\bibitem{Dota2}
Berner, C., Brockman, G., Chan, B., Cheung, V., Debiak, P., Dennison, C., Farhi, D., Fischer, Q., Hashme, S., Hesse, C., J{\'{o}}zefowicz, R., Gray, S., Olsson, C., Pachocki, J., Petrov, M., de~Oliveira~Pinto, H.P., Raiman, J., Salimans, T., Schlatter, J., Schneider, J., Sidor, S., Sutskever, I., Tang, J., Wolski, F., Zhang, S.: Dota 2 with large scale deep reinforcement learning. CoRR  \textbf{abs/1912.06680} (2019), \url{http://arxiv.org/abs/1912.06680}

\bibitem{chen2016net2net}
Chen, T., Goodfellow, I., Shlens, J.: Net2net: Accelerating learning via knowledge transfer (2016)

\bibitem{goodfellow2014generative}
Goodfellow, I.J., Pouget-Abadie, J., Mirza, M., Xu, B., Warde-Farley, D., Ozair, S., Courville, A., Bengio, Y.: Generative adversarial networks (2014)

\bibitem{hinton2015distilling}
Hinton, G., Vinyals, O., Dean, J.: Distilling the knowledge in a neural network (2015)

\bibitem{mobilenet}
Howard, A.G., Zhu, M., Chen, B., Kalenichenko, D., Wang, W., Weyand, T., Andreetto, M., Adam, H.: Mobilenets: Efficient convolutional neural networks for mobile vision applications. CoRR  \textbf{abs/1704.04861} (2017), \url{http://arxiv.org/abs/1704.04861}

\bibitem{neurogenesis}
Kempermann, G., Gage, F.H., Aigner, L., Song, H., Curtis, M.A., Thuret, S., Kuhn, H.G., Jessberger, S., Frankland, P.W., Cameron, H.A., Gould, E., Hen, R., Abrous, D.N., Toni, N., Schinder, A.F., Zhao, X., Lucassen, P.J., Frisén, J.: Human adult neurogenesis: Evidence and remaining questions. Cell Stem Cell  \textbf{23}(1),  25--30 (2018). \doi{https://doi.org/10.1016/j.stem.2018.04.004}, \url{https://www.sciencedirect.com/science/article/pii/S1934590918301668}

\bibitem{CIFAR}
Krizhevsky, A.: Learning multiple layers of features from tiny images (2009), \url{https://api.semanticscholar.org/CorpusID:18268744}

\bibitem{NIPS2012c399862d}
Krizhevsky, A., Sutskever, I., Hinton, G.E.: Imagenet classification with deep convolutional neural networks. In: Pereira, F., Burges, C., Bottou, L., Weinberger, K. (eds.) Advances in Neural Information Processing Systems. vol.~25. Curran Associates, Inc. (2012), \url{https://proceedings.neurips.cc/paper_files/paper/2012/file/c399862d3b9d6b76c8436e924a68c45b-Paper.pdf}

\bibitem{li2020rethinking}
Li, H., Chaudhari, P., Yang, H., Lam, M., Ravichandran, A., Bhotika, R., Soatto, S.: Rethinking the hyperparameters for fine-tuning (2020)

\bibitem{liu2019splitting}
Liu, Q., Wu, L., Wang, D.: Splitting steepest descent for growing neural architectures (2019)

\bibitem{liu2019rethinking}
Liu, Z., Sun, M., Zhou, T., Huang, G., Darrell, T.: Rethinking the value of network pruning (2019)

\bibitem{RonnebergerFB15}
Ronneberger, O., Fischer, P., Brox, T.: U-net: Convolutional networks for biomedical image segmentation. CoRR  \textbf{abs/1505.04597} (2015), \url{http://arxiv.org/abs/1505.04597}

\bibitem{sgd}
Ruder, S.: An overview of gradient descent optimization algorithms. CoRR  \textbf{abs/1609.04747} (2016), \url{http://arxiv.org/abs/1609.04747}

\bibitem{vgg}
Simonyan, K., Zisserman, A.: Very deep convolutional networks for large-scale image recognition (2015)

\bibitem{onpro}
Wei, Y., Ye, J., Huang, Z., Zhang, J., Shan, H.: Online prototype learning for online continual learning (2023)

\bibitem{wu2021firefly}
Wu, L., Liu, B., Stone, P., Liu, Q.: Firefly neural architecture descent: a general approach for growing neural networks (2021)

\end{thebibliography}
\end{document}